\documentclass[conference]{IEEEtran}
\pdfoutput=1
\IEEEoverridecommandlockouts
\usepackage{cite}
\usepackage{amsmath,amssymb,amsfonts}
\usepackage{algorithmic}
\usepackage{graphicx}
\usepackage{textcomp}
\usepackage{xcolor}
\usepackage{subfigure}
\usepackage{multirow}
\usepackage{hyperref}
\def\BibTeX{{\rm B\kern-.05em{\sc i\kern-.025em b}\kern-.08em
    T\kern-.1667em\lower.7ex\hbox{E}\kern-.125emX}}
\begin{document}

\title{A context sensitive real-time Spell Checker with language adaptability}

% \and
% \and
% \and
% \and
% \and
\author{
% \IEEEauthorblockN{Prabhakar Gupta}
% \email{prabhgup@amazon.com}
% \IEEEauthorblockA{\textit{Amazon}}
Prabhakar Gupta \\
Amazon \\
\textit{prabhgup@amazon.com}
}

\maketitle

\begin{abstract}
We present a novel language adaptable spell checking system which detects spelling errors and suggests context sensitive corrections in real-time. We show that our system can be extended to new languages with minimal language-specific processing.Available literature majorly discusses spell checkers for English but there are no publicly available systems which can be extended to work for other languages out of the box. Most of the systems do not work in real-time. We explain the process of generating a language's word dictionary and n-gram probability dictionaries using Wikipedia-articles data and manually curated video subtitles. We present the results of generating a list of suggestions for a misspelled word. We also propose three approaches to create noisy channel datasets of real-world typographic errors. We compare our system with industry-accepted spell checker tools for 11 languages. Finally, we show the performance of our system on synthetic datasets for 24 languages.
\end{abstract}


% \maketitle
\section{Introduction}
Spell checker and correction is a well-known and well-researched problem in Natural Language Processing \cite{Whitelaw2009UsingTW, chen2007improving, gao2010large, Gupta2019ProblemsWA}. However, most state-of-the-art research has been done on spell checkers for English\cite{Choudhury2007HowDI, Dunlop2012MultidimensionalPO}. Some systems might be extended to other languages as well, but there has not been as extensive research in spell checkers for other languages. People have tried to make spell checkers for individual languages: Bengali \cite{islam2007light}, Czech \cite{richter2012korektor}, Danish \cite{bick2006constraint}, Dutch \cite{bosman2013double}, Finnish \cite{pirinen2010finite}, French \cite{starlander2002corpus, Fontenelle2006DevelopingAL}, German \cite{rimrott2008evaluating, Kodydek2000AWA}, Greek \cite{petasis2001greek}, Hindi \cite{etoori2018automatic, Kabra2014AutoSS}, Indonesian \cite{soleh2011non}, Marathi \cite{dixit2005design}, Polish \cite{grundkiewicz2013automatic}, Portuguese \cite{martins1998linguistic}, Russian \cite{sorokin2016automatic, sorokin2017spelling}, Spanish \cite{bustamante2006spelling}, Swedish \cite{kann2001implementation}, Tamil \cite{dhanabalan2003tamil}, Thai \cite{karoonboonyanan1997thai}, etc. This is due to the fact languages are very different in nature and pose different challenges making it difficult to have one solution that work for all languages \cite{Helfrich2000DesignAE}. Many systems do not work in real-time cases. There are some rule-based spell checkers (like LanguageTool\footnote{\url{www.languagetool.org}}) which try to capture grammar and spelling rules \cite{naber2003rule, milkowski2010developing}. This is not scalable and requires language expertise to add new rules. Another problem is evaluating the performance of the spell check system for each language due to lack of quality test data. Spelling errors are classified in two categories \cite{Kukich1992TechniquesFA}: \textit{non-word errors} where the word is unknown and \textit{real-word errors} where the word itself is correct but used in a wrong form / context.

We present a context sensitive real-time spell-checker system which can be adapted to any language. One of the biggest problem earlier was absence of data for languages other than English, so we propose three approaches to create noisy channel datasets of real-world typographic errors. We use Wikipedia data for creating dictionaries and synthesizing test data. To compensate for resource-scarcity of most languages we also use manually curated movie subtitles since it provides information about how people communicate as shown in \cite{prabhakar2019Unsupervised}.

Our system outperforms industry-wide accepted English spell checkers (Hunspell and Aspell) and show our performance on benchmark datasets for English. We present our performance on synthetic dataset for 24 languages \textit{viz.,} Bengali, Czech, Danish, Dutch, English, Finnish, French, German, Greek, Hebrew, Hindi, Indonesian, Italian, Marathi, Polish, Portuguese, Romanian, Russian, Spanish, Swedish, Tamil, Telugu, Thai and Turkish. We also compare 11 of these languages to one of the most popular rule-based systems.
% Our system generates all the possible candidates for a misspelled word in edit-distance 2, ranks them based on the context they are used in, and produces a list of top $k$ results. 
We did not customize our spell checker to suit local variants or dialects of a language. For example --- the spelling \textit{``color''} is used in American English whereas spelling \textit{``colour''} is preferred in other versions of English. Our system will not flag any of these spellings.

The paper makes following contributions:
\begin{itemize}
    % \item We develop a context aware spell checker for 24 languages with extensive experimentation on synthetic datasets.
    \item We propose three different approaches to create typographic errors for any language which has never been done in multilingual setting (all earlier approaches have either been very simple \cite{etoori2018automatic} or language-specific \cite{dixit2005design}).
    \item We show system's time performance for each step in process, proving it's real-time effectiveness. 
    % \item We show effectiveness of the system on the three aforementioned algorithms, on edit distances 1 and 2 and for false-positive and false-negative experiments. 
    \item Our system outperforms existing rule-based and industry-wide accepted spell checking tools. 
    \item We show that our system can be adapted to other languages with minimal effort --- showing precision@k for $k \in {1, 3, 5, 10}$ and mean reciprocal rank (MRR) for 24 languages.
    % \item We present the results of our experiment to decide the maximum n-grams to be considered.
\end{itemize}
% with three different  
% Spelling errors are classified in broadly two categories: \textit{real-world errors (RWE)} and \textit{non-words error (NWE)}. We focus on NWE in this paper.
% The biggest problem was evaluating the accuracy of the system for all languages, therefore, we also present the two algorithms we used in order to synthesize spelling errors. 

The paper is divided into four sections. Section \ref{section:approach} explains the preprocessing steps and approach to generate a ranked list of suggestions for any detected error. Section \ref{section:synthesizespellingerror} presents different synthetic data-generation algorithms. 
% Section \ref{section:experiment} describes evaluation experiment descriptions and their results which include results on synthetic data, comparison, to use for ranking suggestions. 
Section \ref{section:experiment} describes the experiments and reports their results.
Finally, Section \ref{section:conclusion} concludes the paper and discusses future endeavours.

\section{Approach}
\label{section:approach}
Our system takes a sentence as input, tokenizes the sentence, identifies misspelled words (if any), generates a list of suggestions and ranks them to return the top $k$ corrections. For ranking the suggestions, we use n-gram conditional probabilities. As a preprocessing step, we create frequency dictionaries which will aid in generation of n-gram conditional probabilities.

\subsection{Preprocessing: Building n-gram dictionaries}
We calculated unigram, bigram and trigram frequencies of tokens from corpus. Using these frequencies, we calculated conditional probabilities expressed in the equation \ref{equation:conditionalprob} where $P$ is conditional probability and $c$ is the count of the n-gram in corpus. For unigrams, we calculate its probability of occurrence in the corpus.

\begin{equation}
\label{equation:conditionalprob}
P(w_{i} | w_{i-n+1} ... w_{i-1}) = \frac{c(w_{i-n+1} ... w_{i})}{c(w_{i-n+1} ... w_{i-1})}   
\end{equation}

We used Wikipedia dumps\footnote{Wikimedia Downloads: \url{https://dumps.wikimedia.org}} along with manually curated movie subtitles for all languages. We capped Wikipedia articles to 1 million and subtitle files to 10K. On an average, each subtitle file contains 688 subtitle blocks and each block contains 6.4 words \cite{prabhakar2019Unsupervised}. 
% As shown in Table \ref{table:dataanalysis}, popular languages like English, German, French, etc. have plenty of Wikipedia articles while regional languages like Marathi, Bengali, Telugu, Hindi, Tamil and Thai have less than $150k$ Wikipedia articles. 
We considered words of minimum length 2 with frequency more than 5 times in the corpus. Similarly, only bigrams and trigrams where each token was known were considered.

% \begin{table}[h]
%   \centering
%   \caption{Preprocessing Data Analysis}
%   \label{table:dataanalysis}
% %   \resizebox{\columnwidth}{!}{%
%     \begin{tabular}{l|rrr}
%     \centering
%     \multirow{2}{*}{\textbf{Language}} & \textbf{\# Wikipedia} &  \textbf{\# Subtitle} & \textbf{Dictionary} \\ 
%     & \textbf{articles} & \textbf{blocks} & \textbf{Size} \\ \hline
%     % \textbf{} & \textbf{ } & \textbf{ } & \textbf{} \\ \hline
%         Bengali & 64.9K & 824.3K & 130,302 \\
%         Czech & 422.9K & 50.1K & 501,007 \\
%         Danish & 243.3K & 1.1M & 305,414 \\
%         Dutch & 1.1M & 3.8M & 506,969 \\
%         English & 1.0M & 11.9M & 633,202 \\
%         Finnish & 452.4K & 486.4K & 705,209 \\
%         French & 1.0M & 4.5M & 595,836 \\
%         German & 1.0M & 5.9M & 1,213,171 \\
%         Greek & 159.0K & 48.5K & 111,878 \\
%         Hebrew & 238.7K & 53.9K & 378,343 \\
%         Hindi & 125.6K & 2.8M & 123,348 \\
%         Indonesian & 453.9K & 1.0M & 274,707 \\
%         Italian & 1.1M & 4.4M & 594,512 \\
%         Marathi & 53.5K & 807.7K & 97,099 \\
%         Polish & 1.1M & 4.0M & 840,283 \\
%         Portuguese & 1.0M & 6.8M & 536,855 \\
%         Romanian & 391.4K & 57.1K & 271,190 \\
%         Russian & 1.0M & 3.6M & 1,055,876 \\
%         Spanish & 1.0M & 6.3M & 669,246 \\
%         Swedish & 1.2M & 1.1M & 571,690 \\
%         Tamil & 126.7K & 332.5K & 223,988 \\
%         Telugu & 70.5K & 1.1M & 211,194 \\
%         Thai & 130.2K & 25.5K & 262,805 \\
%         Turkish & 323.7K & 4.3M & 458,083 \\
%     \end{tabular}
%     % }
% \end{table}

One issue we encountered while building these dictionaries using such a huge corpus was its size. For English, the number of unique unigrams was approx. $2.2M$, bigrams was $50M$ and trigrams was $166M$. If we store these files as uncompressed Python Counters, these files end up being $44MB$, $1.8GB$ and $6.4GB$ respectively. To reduce the size, we compressed these files using a word-level Trie with hashing. We created a hash map for all the words in the dictionary (unigram token frequency) assigning a unique integer id to each word. Using each word's id, we created a trie-like structure where each node represented one id and its children represented n-grams starting with that node's value. The Trie ensured that the operation to lookup an n-gram was bounded in \textit{O(1)} and reduced the size of files by $66\%$ on an average. For English, the hashmap was $14MB$, unigram probabilities' file was $8.7MB$, bigram was $615MB$ and trigram was $2.5GB$.

\subsection{Tokenization}
There are a number of solutions available for creating tokenizer for multiple languages. Some solutions (like \cite{Moreau2018MultilingualWS, Snyder2008UnsupervisedML, Chang2008OptimizingCW}), try to use publicly available data to train tokenizers, whereas some solutions (like Europarl preprocessing tools \cite{Koehn2005EuroparlAP}) are rule-based. Both approaches are not extensible and typically are not real-time. 
% Our goal is to create a \textit{simpler} solution which can be adapted to new languages with minimal language expertise. 

For a language, we create list of supported characters using writing systems information\footnote{\url{https://en.wikipedia.org/wiki/List_of_languages_by_writing_system}} and Language recognition charts\footnote{\url{https://en.wikipedia.org/wiki/Wikipedia:Language_recognition_chart}}. We included uppercase and lowercase characters (if applicable) and numbers in that writing system, ignoring all punctuation. Any character which doesn't belong to this list is implied as foreign character to that language and will be tokenized as a separate token. Using regex rule, we extract all continuous sequences of characters in supported list.
% We used different tokenizers for different sets of languages as suggested in \cite{bojanowski2016enriching}. For Latin, Cyrillic, Hebrew and Greek scripts, we use ; for Arabic, we use pyarabic library \cite{zerrouki2012pyarabic}; for Chinese, we use Stanford Word Segmenter \cite{} 

\subsection{Error Detection}
We kept our error-search strictly to non-words errors; for every token in sentence, we checked for its occurrence in dictionary. However, to make system more efficient, we only considered misspelled tokens of length greater than 2. On manual analysis of Wikipedia misspellings dataset for English, we discovered misspelling of length 1 and 2 do not make sense and hence computing suggestions and ranking them is not logical.

\subsection{Generating candidate suggestions}
Given an unknown token, we generated a list of all known words within edit distance of 2, calling them candidate suggestions. We present the edit distance distribution of publicly available datasets for English in Section \ref{section:publicdatasetsresults}. Two intuitive approaches to generate the list of suggestions that work fairly well on a small-size dataset are checking edit-distance of incorrect spelling with all words in dictionary and second, generating a list of all words in edit-distance 2 of incorrect spelling\footnote{\url{https://norvig.com/spell-correct.html}}. The obvious problem with the first approach is with the size of corpus which is typically in range of hundreds of thousands and with the second approach is size of word because for longer words there can be thousands of suggestions and building a list of such words is also time consuming.

We considered four approaches --- Trie data structure, Burkhard-Keller Tree (BK Tree) \cite{Burkhard1973SomeAT}, Directed Acyclic Word Graphs (DAWGs) \cite{Balk2002ImplementationOD} and Symmetric Delete algorithm (SDA)\footnote{\url{https://github.com/wolfgarbe/SymSpell}}. 
% Figure \ref{fig:suggestionperformance-1} represents the performance of four algorithms for edit distance 1 and Figure \ref{fig:suggestionperformance-2} represents the performance on edit distance 2. 
In Table \ref{table:suggestionperformance-2}, we represent the performance of algorithms for edit distance 2 without adding results for BK trees because its performance was in range of couple of seconds. We used Wikipedia misspelling dataset\footnote{\url{https://en.wikipedia.org/wiki/Wikipedia:Lists_of_common_misspellings}} to create a list of 2062 unique misspellings of lengths varying from 3 to 16 which were not present in our English dictionary. For each algorithm, we extracted the list of suggestions in edit distance of 1 and 2 for each token in dataset. 
% SDA outperforms all other available approaches. It was also better than other approaches in terms of space. For English, final SDA Python pickle object took $130MB$ whereas Trie took $230MB$.

% \begin{figure*}[t]
% \centering
% \subfigure[Edit Distance = 1]{
% \includegraphics[width=.45\linewidth]{images/timePerformanceEditDistance_1.pdf}
% \label{fig:suggestionperformance-1}
% }
% \subfigure[Edit Distance = 2]{
% \includegraphics[width=.45\linewidth]{images/timePerformanceEditDistance_2.pdf}
% \label{fig:suggestionperformance-2}
% }
% \caption{Time Performance of suggestion generation algorithms}
% \end{figure*}

% \begin{figure}[h]
%     \includegraphics[width=\linewidth]{images/timePerformanceEditDistance_1.pdf}
%     \caption{Time Performance of suggestion generation algorithms (Edit Distance = 1)}
%     \label{fig:suggestionperformance-1}
% \end{figure}

% \begin{figure}[h]
%     \includegraphics[width=\linewidth]{images/timePerformanceEditDistance_2.pdf}
%     \caption{Time Performance of suggestion generation algorithms (Edit Distance = 2)}
%     \label{fig:suggestionperformance-2}
% \end{figure}

\begin{table}[h]
  \centering
  \caption{Average Time taken by suggestion generation algorithms (Edit Distance = 2) (in millisecond)}
  \label{table:suggestionperformance-2}
%   \resizebox{\columnwidth}{!}{%
    \begin{tabular}{c|rrr}
    \centering
    \textbf{Token} & \textbf{Trie} & \textbf{DAWGs} & \textbf{SDA} \\ \hline
        3 & 170.50 & 180.98 & 112.31 \\
        4 & 175.04 & 178.78 & 52.97 \\
        5 & 220.44 & 225.10 & 25.44 \\
        6 & 254.57 & 259.54 & 7.44 \\
        7 & 287.19 & 291.99 & 4.59 \\
        8 & 315.78 & 321.58 & 2.58 \\
        9 & 351.19 & 356.76 & 1.91 \\
        10 & 379.99 & 386.04 & 1.26 \\
        11 & 412.02 & 419.55 & 1.18 \\
        12 & 436.54 & 443.85 & 1.06 \\
        13 & 473.45 & 480.26 & 1.16 \\
        14 & 508.08 & 515.04 & 0.97 \\
        15 & 548.04 & 553.49 & 0.66 \\
        16 & 580.44 & 584.99 & 0.37 \\
    \end{tabular}
    % }
\end{table}

% We considered a Trie data-structure based appraoch \cite{Burkhard1973SomeAT}. We build a trie for each language using all the valid words in that language. We define a word to be valid if it has frequency of atleast 5 in corpus.
% To get suggestion, we search the data structure maintaining a counter: \textit{maximum allowed cost} to enforce we only get words which have edit distance 2. We present the time performance of this approach in Appendix \ref{appendix:trietimeperformance}

\subsection{Ranking suggestions}
Using SDA, we generate a list of candidates which are to be ranked in order of relevance in the given context. Authors of \cite{Carlson2007MemorybasedCS}, demonstrate the effectiveness of n-grams for English to auto-correct real-word errors and unknown word errors. However, they use high-order n-grams in isolation. We propose a weighted sum of unigrams, bigrams and trigrams to rank the suggestions. 
% We show in Section \ref{section:decidingmaxn-gram}, why using n-grams for $n > 3$ is not computationally optimum. 
Authors in \cite{Fivez2017UnsupervisedCS}, use character embeddings to generate embeddings for each misspelling for clinical free-text and then similar to \cite{Kilicoglu2015AnEM}, rank on basis of contextual similarity score.

We create a context score ($S$) for each suggestion and rank on decreasing order of that score, returning top $k$ suggestions. Context score is weighted sum of unigram context score ($S_1$), bigram context score ($S_2$) and trigram context score ($S_3$) defined by equation \ref{equation:scoringngram}. This score is calculated for each suggestion by replacing token $x_i$ with the suggestion. For n-grams where any token is unknown, the count is considered to be $0$.

\begin{equation}
\label{equation:scoringngram}
     S_{n} = W_{n} \sum_{j=0}^{n-1} \frac{c(x_{i+j-n+1}^{i+j})}{c(x_{i+j-n+1}^{i+j-1})} = W_{n} \sum_{j=0}^{n-1} {P(x_i | x_{i+j-n+1}^{i+j-1})}
\end{equation}
where:

$i$ = index of misspelled token

$W_n$ = the weight for $n^{th}$-gram's score

$c(x_i^j)$ = occurrence frequency of sequence ($w_i$ \ldots $w_j$)

$P$ = conditional probability.

% We experimented with a bunch of approaches including word embeddings  and ngrams. We explain both the approaches and show the results for ngram approach as the results of word embeddings approach were not promising.

% $$
% \fbox{x_{1}} \fbox{x_{2}} \fbox{\ldots} \fbox{x_{i-2}} \fbox{x_{i-1}} \fbox{x_{i}} \fbox{x_{i+1}} \fbox{x_{i+2}} \fbox{\ldots} \fbox{a}
% $$

% \begin{equation}
% \label{equation:ngramcalculation}
%     w_{1}[p_{1}(X)] 
% \end{equation}

% \begin{equation}
% \label{equation:ngramcalculation}
%     w_{2} \left [p_{2}(x_{i-1}, X) + p_{2}(X, x_{i+1}) \right ]
% \end{equation}

% \begin{equation}
% \label{equation:ngramcalculation}
%     w_{3} \left [p_{3}(x_{i-2}, x_{i-1}, X) + p_{3}(x_{i-1}, X, x_{i+1}) + p_{3}(X, x_{i+1}, x_{i+2}) \right ] 
% \end{equation}

\section{Synthesizing Spelling Errors}
\label{section:synthesizespellingerror}
The biggest challenge in evaluation of spell checker was quality test dataset. Most of the publicly available datasets are for English \cite{wiked_2014}. We propose three strategies to introduce typographical errors in correct words to represent noisy channel. We select all the sentences, where we did not find any spelling error and introduced exactly one error per sentence.

\subsection{Randomized Characters}
From a sentence, we pick one word at random and make one of the three edits: insertion, deletion or substitution with a random character from that language's supported character list. Since it is a completely randomized strategy, the incorrect words created are not very ``realistic''. For example --- in English for edit distance 2, word \textit{``moving''} was changed to \textit{``moviAX''}, \textit{``your''} to \textit{``mouk''}, \textit{``chest''} to \textit{``chxwt''}. We repeated the process for edit distance 1 (introducing only one error) and edit distance 2 (introducing two errors) and create dataset for 20,000 sentences each.

\begin{figure*}[t]
\centering
\subfigure[Variation of unigram weight ($W_1$)]{
\includegraphics[width=.31\linewidth]{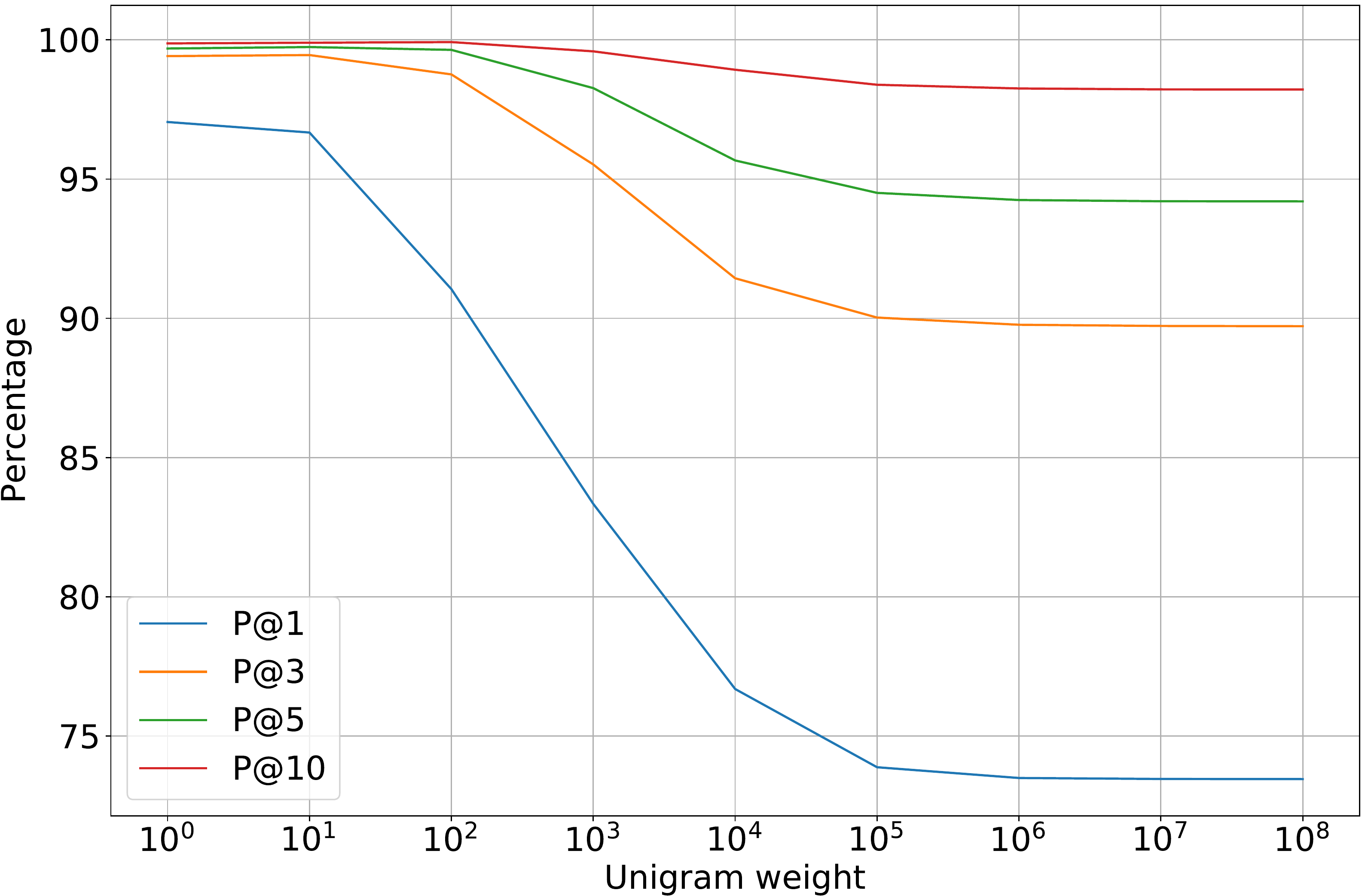}
\label{fig:ngramDependency1}
}
\subfigure[Variation of bigram weight ($W_2$)]{
\includegraphics[width=.31\linewidth]{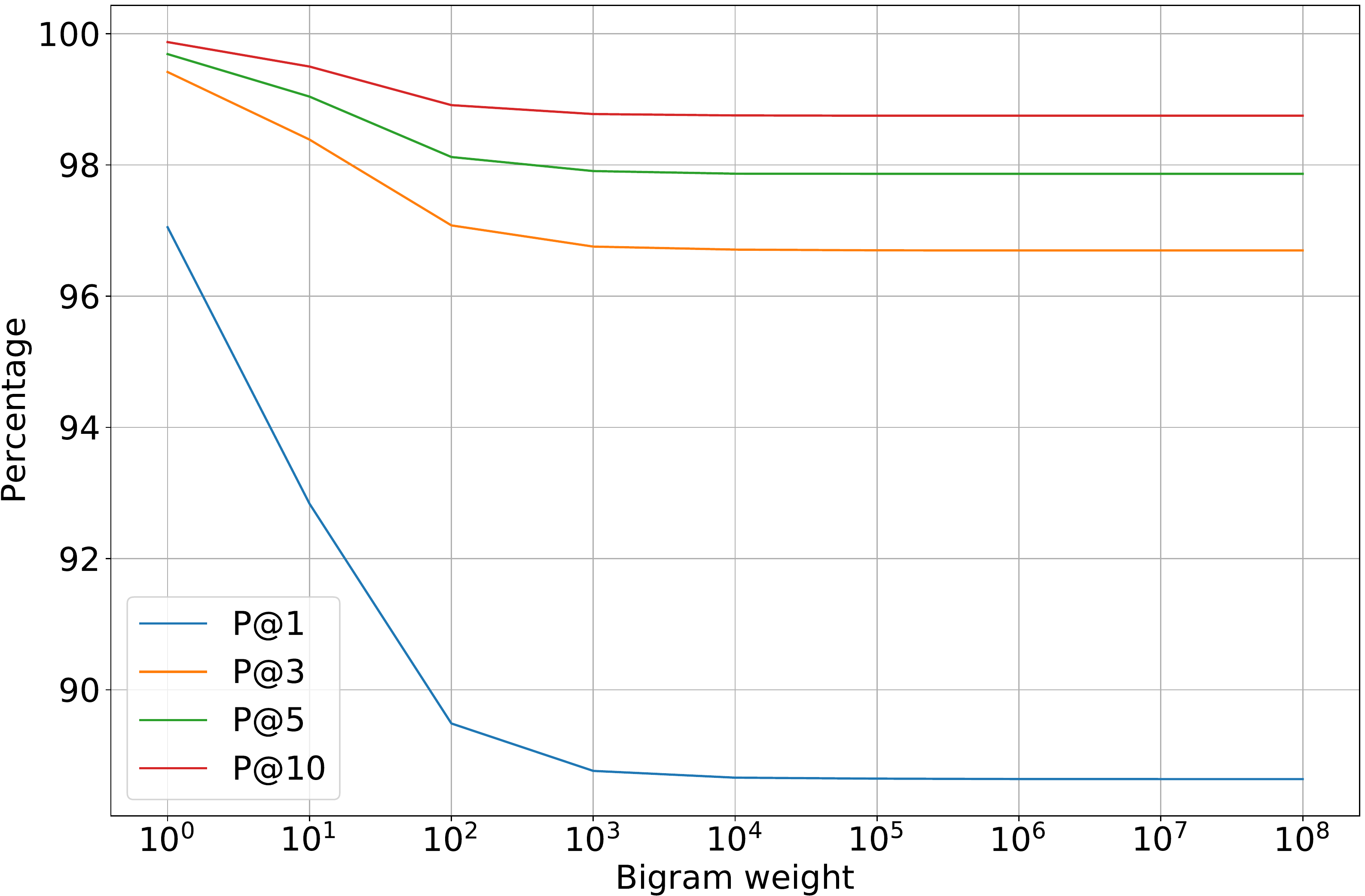}
\label{fig:ngramDependency2}
}
\subfigure[Variation of trigram weight ($W_3$)]{
\includegraphics[width=.31\linewidth]{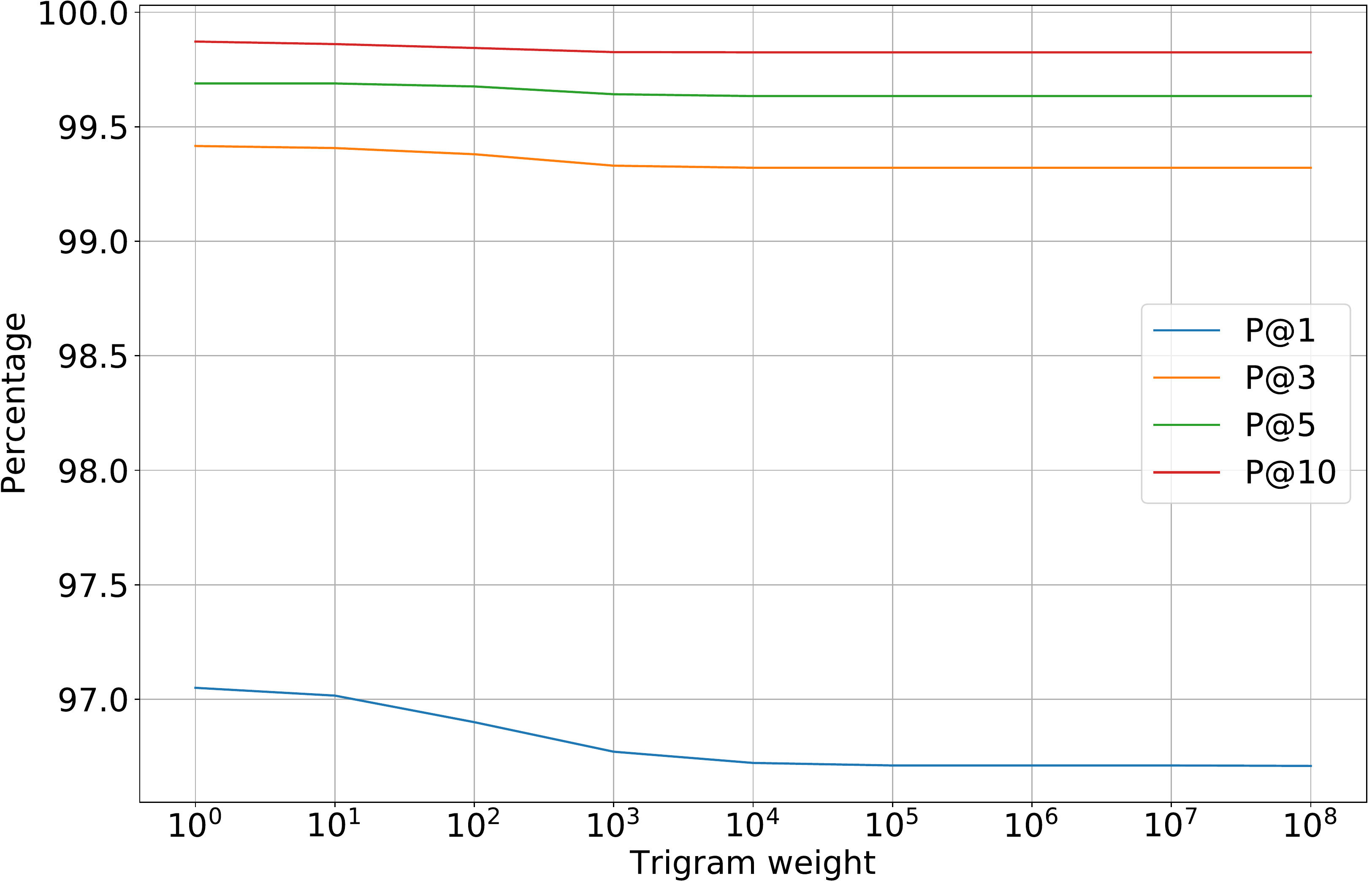}
\label{fig:ngramDependency3}
}

\caption{Importance of n-grams weights towards system accuracy}
\label{fig:ngramDependency}
\end{figure*}

\subsection{Characters Swap}
On analyzing common misspellings for English\cite{wiked_2014}, we discovered majority of edit-distance 2 errors are swap of two adjacent characters. For example --- \textit{``grow''} is misspelled as \textit{``gorw''}, \textit{``grief''} as \textit{``greif''}. One swap imply edit distance of two, we created a dataset of 20,000 samples for such cases.

\subsection{Character Bigrams}
Introducing errors randomly produces \textit{unrealistic} words. To create more realistic errors, we decided to use character bigram information. From all the words in dictionary for a language, we calculate occurrence probabilities for character bigrams.
For a given word, we select a character bigram randomly and replace the second character in selected bigram with a possible substitute from pre-computed character bigram probabilities. This way, we were able to generate words which were more plausible. For example --- in English for edit distance 1, word \textit{``heels''} was changed to \textit{``heely''}, \textit{``triangle''} to \textit{``triajgle''}, \textit{``knee''} to \textit{``kyee''}. On shallow manual analysis of generated words, most of the words look quite realistic. For English, some of the words generated are representative of keyboard-strokes error (errors that occur due to mistakenly pressing a near-by key on keyboard/input device). For example, we generated some samples like --- \textit{``Allow''} to \textit{``Alkow''}, \textit{``right''} to \textit{``riggt''}, \textit{``flow''} to \textit{``foow''} and \textit{``Stand''} to \textit{``Stabd''}.  We generated a sample of 40,000 sentences each for edit distance 1 and edit distance 2.

\section{Experiments and Results}
\label{section:experiment}
\subsection{Synthetic Data evaluation}
For each language, we created a dataset of 140,000 sentences\footnote{With an exception of Czech, Greek, Hebrew and Thai where size of dataset was smaller due to unavailability of good samples} with one misspelling each. 
% For hyperparameter optimization (HPO), we used Random Search\cite{Bergstra2012RandomSF}. We present the performance for different values of weights (sorted with $P@1$) for English in Table \ref{table:hpo}. 
The best performances for each language is reported in Table \ref{table:syntheticdataresults}. We present Precision@k\footnote{Percentage of cases where expected output was in top $k$ results} for $k \in {1, 3, 5, 10}$ and mean reciprocal rank (MRR). The system performs well on synthetic dataset with a minimum of 80\% \textit{P@1} and 98\% \textit{P@10}.

\begin{table}[h]
  \centering
  \caption{Synthetic Data Performance results}
  \label{table:syntheticdataresults}
  \resizebox{\columnwidth}{!}{%
    \begin{tabular}{l|c|ccccc}
    \centering
    \multirow{2}{*}{\textbf{Language}} & \textbf{\# Test} & \multirow{2}{*}{\textbf{P@1}} & \multirow{2}{*}{\textbf{P@3}} & \multirow{2}{*}{\textbf{P@5}} & \multirow{2}{*}{\textbf{P@10}} & \multirow{2}{*}{\textbf{MRR}} \\ 
     & \textbf{Samples} & & & &  \\ \hline
    % \textbf{} & \textbf{ } & \textbf{P@1} & \textbf{P@3} & \textbf{P@5} & \textbf{P@10} & \textbf{MRR} \\ \hline
        Bengali & 140000 & 91.30 & 97.83 & 98.94 & 99.65 & 94.68 \\
        Czech & 94205 & 95.84 & 98.72 & 99.26 & 99.62 & 97.37 \\
        Danish & 140000 & 85.84 & 95.19 & 97.28 & 98.83 & 90.85 \\
        Dutch & 140000 & 86.83 & 95.01 & 97.04 & 98.68 & 91.32 \\
        English & 140000 & 97.08 & 99.39 & 99.67 & 99.86 & 98.27 \\
        Finnish & 140000 & 97.77 & 99.58 & 99.79 & 99.90 & 98.69 \\
        French & 140000 & 86.52 & 95.66 & 97.52 & 98.83 & 91.38 \\
        German & 140000 & 87.58 & 96.16 & 97.86 & 99.05 & 92.10 \\
        Greek & 30022 & 84.95 & 94.99 & 96.88 & 98.44 & 90.27 \\
        Hebrew & 132596 & 94.00 & 98.26 & 99.05 & 99.62 & 96.24 \\
        Hindi & 140000 & 82.19 & 93.71 & 96.28 & 98.30 & 88.40 \\
        Indonesian & 140000 & 95.01 & 98.98 & 99.50 & 99.84 & 97.04 \\
        Italian & 140000 & 89.93 & 97.31 & 98.54 & 99.38 & 93.76 \\
        Marathi & 140000 & 93.01 & 98.16 & 99.06 & 99.66 & 95.69 \\
        Polish & 140000 & 95.65 & 99.17 & 99.62 & 99.86 & 97.44 \\
        Portuguese & 140000 & 86.73 & 96.29 & 97.94 & 99.10 & 91.74 \\
        Romanian & 140000 & 95.52 & 98.79 & 99.32 & 99.68 & 97.22 \\
        Russian & 140000 & 94.85 & 98.74 & 99.33 & 99.71 & 96.86 \\
        Spanish & 140000 & 85.91 & 95.35 & 97.18 & 98.57 & 90.92 \\
        Swedish & 140000 & 88.86 & 96.40 & 98.00 & 99.14 & 92.87 \\
        Tamil & 140000 & 98.05 & 99.70 & 99.88 & 99.98 & 98.88 \\
        Telugu & 140000 & 97.11 & 99.68 & 99.92 & 99.99 & 98.38 \\
        Thai & 12403 & 98.73 & 99.71 & 99.78 & 99.85 & 99.22 \\
        Turkish & 140000 & 97.13 & 99.51 & 99.78 & 99.92 & 98.33 \\
    \end{tabular}
    }
\end{table}

\begin{table}[h]
  \centering
  \caption{Synthetic Data Time Performance results}
  \label{table:syntheticdatatimeresults}
  \resizebox{\columnwidth}{!}{%
    \begin{tabular}{l|c|cc|c}
    \centering
    \multirow{2}{*}{\textbf{Language}} & \textbf{Detection} &  \multicolumn{2}{|c|}{\textbf{Suggestion Time}} & \textbf{Ranking} \\ 
    & \textbf{Time ($\mu$s)} & \textbf{ED=1 (ms)} & \textbf{ED=2 (ms)} & \textbf{Time (ms)} \\ \hline
Bengali & 7.20 & 0.48 & 14.85 & 1.14 \\
Czech & 7.81 & 0.75 & 26.67 & 2.34 \\
Danish & 7.28 & 0.67 & 23.70 & 1.96 \\
Dutch & 10.80 & 0.81 & 30.44 & 2.40 \\
English & 7.27 & 0.79 & 39.36 & 2.35 \\
Finnish & 8.53 & 0.46 & 15.55 & 1.05 \\
French & 7.19 & 0.82 & 32.02 & 2.69 \\
German & 8.65 & 0.85 & 41.18 & 2.63 \\
Greek & 7.63 & 0.86 & 25.40 & 1.87 \\
Hebrew & 22.35 & 1.01 & 49.91 & 2.18 \\
Hindi & 8.50 & 0.60 & 18.51 & 1.72 \\
Indonesian & 12.00 & 0.49 & 20.75 & 1.22 \\
Italian & 6.92 & 0.72 & 29.02 & 2.17 \\
Marathi & 7.16 & 0.43 & 10.68 & 0.97 \\
Polish & 6.44 & 0.64 & 24.15 & 1.74 \\
Portuguese & 7.14 & 0.66 & 28.92 & 2.20 \\
Romanian & 10.26 & 0.63 & 18.83 & 1.79 \\
Russian & 6.79 & 0.68 & 22.56 & 1.72 \\
Spanish & 7.19 & 0.75 & 31.00 & 2.41 \\
Swedish & 7.76 & 0.83 & 32.17 & 2.57 \\
Tamil & 11.34 & 0.23 & 4.83 & 0.31 \\
Telugu & 6.31 & 0.29 & 7.50 & 0.54 \\
Thai & 11.60 & 0.66 & 18.75 & 1.33 \\
Turkish & 7.40 & 0.49 & 17.42 & 1.23 \\
    \end{tabular}
    }
\end{table}

The system is able to do each sub-step in real-time; the average time taken to perform for each sub-step is reported in Table \ref{table:syntheticdatatimeresults}. All the sentences used for this analysis had exactly one error according to our system. Detection time is the average time weighted over number of tokens in query sentence, suggestion time is weighted over misspelling character length and ranking time is weighted over length of suggestions generated.

Table \ref{table:syntheticdataalgorithmresults} presents the system's performance on each error generation algorithm. We included only \textit{P@1} and \textit{P@10} to show trend on all languages. ``Random Character'' and ``Character Bigrams'' includes data for edit distance 1 and 2 whereas ``Characters Swap'' includes data for edit distance 2. Table \ref{table:syntheticdataeditdistanceresults} presents the system's performance individually on edit distance 1 and 2. We included only \textit{P@1}, \textit{P@3} and \textit{P@10} to show trend on all languages.

\begin{table}[h]
  \centering
  \caption{Synthetic Data Performance on three error generation algorithm}
  \label{table:syntheticdataalgorithmresults}
  \resizebox{\columnwidth}{!}{%
    \begin{tabular}{l|cc|cc|cc}
    \centering
    %  & \textbf{\# Test} & \multirow{2}{*}{\textbf{P@1}} & \multirow{2}{*}{\textbf{P@3}} & \multirow{2}{*}{\textbf{P@5}} & \multirow{2}{*}{\textbf{P@10}} & \multirow{2}{*}{\textbf{MRR}} \\ 
    %  & \textbf{Samples} & & & &  \\ \hline
     \multirow{2}{*}{\textbf{Language}} & \multicolumn{2}{c|}{\textbf{Random Character}} & \multicolumn{2}{c|}{\textbf{Characters Swap}} & \multicolumn{2}{c}{\textbf{Character Bigrams}} \\ \cline{2-7}
     & \textbf{P@1} & \textbf{P@10} & \textbf{P@1} & \textbf{P@10} & \textbf{P@1} & \textbf{P@10} \\ \hline
        Bengali & 91.243 & 99.493 & 82.580 & 99.170 & 93.694 & 99.865 \\
        Czech & 94.035 & 99.264 & 91.560 & 99.154 & 97.795 & 99.909 \\
        Danish & 84.605 & 98.435 & 71.805 & 97.160 & 90.103 & 99.444 \\
        Dutch & 85.332 & 98.448 & 72.800 & 96.675 & 91.159 & 99.305 \\
        English & 97.260 & 99.897 & 93.220 & 99.700 & 98.050 & 99.884 \\
        Finnish & 97.735 & 99.855 & 94.510 & 99.685 & 98.681 & 99.972 \\
        French & 84.332 & 98.483 & 72.570 & 97.215 & 91.165 & 99.412 \\
        German & 86.870 & 98.882 & 73.920 & 97.550 & 91.448 & 99.509 \\
        Greek & 82.549 & 97.800 & 71.925 & 96.910 & 90.291 & 99.386 \\
        Hebrew & 94.180 & 99.672 & 88.491 & 99.201 & 95.414 & 99.706 \\
        Hindi & 81.610 & 97.638 & 67.730 & 96.200 & 86.274 & 99.169 \\
        Indonesian & 94.735 & 99.838 & 89.035 & 99.560 & 96.745 & 99.910 \\
        Italian & 88.865 & 99.142 & 78.765 & 98.270 & 93.400 & 99.775 \\
        Marathi & 92.392 & 99.493 & 85.145 & 99.025 & 95.449 & 99.905 \\
        Polish & 94.918 & 99.743 & 90.280 & 99.705 & 97.454 & 99.954 \\
        Portuguese & 86.422 & 98.903 & 71.735 & 97.685 & 90.787 & 99.562 \\
        Romanian & 94.925 & 99.575 & 90.805 & 99.245 & 97.119 & 99.845 \\
        Russian & 93.285 & 99.502 & 89.000 & 99.240 & 97.196 & 99.942 \\
        Spanish & 84.535 & 98.210 & 71.345 & 96.645 & 90.395 & 99.246 \\
        Swedish & 87.195 & 98.865 & 76.940 & 97.645 & 92.828 & 99.656 \\
        Tamil & 98.118 & 99.990 & 96.920 & 99.990 & 99.284 & 99.999 \\
        Telugu & 97.323 & 99.990 & 93.935 & 99.985 & 97.897 & 99.998 \\
        Thai & 97.989 & 99.755 & 97.238 & 99.448 & 98.859 & 99.986 \\
        Turkish & 97.045 & 99.880 & 93.195 & 99.815 & 98.257 & 99.972 \\
    \end{tabular}
    }
\end{table}

We experimented with the importance of each n-gram. Figure \ref{fig:ngramDependency} presents the results for this experiment. We kept two weights constant varying one weight to compare the performance. For example to determine unigram weight ($W_1$) importance, we set bigram weight ($W_2$) and trigram ($W_3$) to $1$, varying $W_1$ ($10^i, i \in [0,8]$). As shown in Figure \ref{fig:ngramDependency1} and Figure \ref{fig:ngramDependency2}, if unigram or trigram are given more importance, the performance of system worsens. Figure \ref{fig:ngramDependency3} shows removing lower order n-grams and giving more importance to only trigram also decreases performance. Therefore, finding the right balance between each weight is crucial for system's best performance.

\begin{table}[h]
  \centering
  \caption{Synthetic Data Performance on different edit distance of errors}
  \label{table:syntheticdataeditdistanceresults}
  \resizebox{\columnwidth}{!}{%
    \begin{tabular}{l|ccc|ccc}
    \centering
    %  & \textbf{\# Test} & \multirow{2}{*}{\textbf{P@1}} & \multirow{2}{*}{\textbf{P@3}} & \multirow{2}{*}{\textbf{P@5}} & \multirow{2}{*}{\textbf{P@10}} & \multirow{2}{*}{\textbf{MRR}} \\ 
    %  & \textbf{Samples} & & & &  \\ \hline
     \multirow{2}{*}{\textbf{Language}} & \multicolumn{3}{c|}{\textbf{Edit Distance = 1}} &  \multicolumn{3}{c}{\textbf{Edit Distance = 2}} \\ \cline{2-7}
     & \textbf{P@1} & \textbf{P@3} & \textbf{P@10} & \textbf{P@1} & \textbf{P@3} & \textbf{P@10} \\ \hline
        Bengali & 97.475 & 99.883 & 99.998 & 86.581 & 96.282 & 99.395 \\
        Czech & 98.882 & 99.914 & 99.996 & 93.016 & 97.611 & 99.271 \\
        Danish & 95.947 & 99.692 & 99.970 & 78.272 & 91.797 & 97.960 \\
        Dutch & 96.242 & 99.653 & 99.958 & 79.790 & 91.528 & 97.722 \\
        English & 99.340 & 99.985 & 99.998 & 95.400 & 98.954 & 99.750 \\
        Finnish & 99.398 & 99.968 & 99.998 & 96.549 & 99.280 & 99.820 \\
        French & 95.645 & 99.658 & 99.985 & 79.706 & 92.664 & 97.959 \\
        German & 96.557 & 99.807 & 99.983 & 80.866 & 93.431 & 98.345 \\
        Greek & 94.964 & 99.538 & 99.964 & 76.102 & 90.980 & 97.096 \\
        Hebrew & 97.643 & 99.715 & 99.990 & 90.217 & 96.883 & 99.313 \\
        Hindi & 93.127 & 99.590 & 99.997 & 73.731 & 89.276 & 97.025 \\
        Indonesian & 98.687 & 99.955 & 99.995 & 92.091 & 98.231 & 99.716 \\
        Italian & 95.818 & 99.670 & 99.978 & 84.585 & 95.370 & 98.912 \\
        Marathi & 96.262 & 99.700 & 99.993 & 89.524 & 96.834 & 99.401 \\
        Polish & 96.925 & 99.728 & 99.997 & 93.246 & 98.585 & 99.749 \\
        Portuguese & 95.903 & 99.872 & 99.995 & 79.889 & 93.597 & 98.436 \\
        Romanian & 98.690 & 99.897 & 99.988 & 93.156 & 97.942 & 99.439 \\
        Russian & 97.568 & 99.830 & 99.992 & 92.257 & 97.851 & 99.499 \\
        Spanish & 95.190 & 99.627 & 99.977 & 78.950 & 92.140 & 97.520 \\
        Swedish & 96.932 & 99.778 & 99.968 & 82.836 & 93.865 & 98.511 \\
        Tamil & 97.120 & 99.873 & 99.998 & 98.204 & 99.808 & 99.996 \\
        Telugu & 95.985 & 99.853 & 99.998 & 95.662 & 99.445 & 99.989 \\
        Thai & 96.994 & 99.470 & 99.983 & 97.786 & 99.450 & 99.725 \\
        Turkish & 98.635 & 99.927 & 99.998 & 95.521 & 99.164 & 99.865 \\
    \end{tabular}
    }
\end{table}

% \begin{figure}[!htb]
% \subfloat[]{%
%   \includegraphics[width=\columnwidth]{images/ngramDependency_1.pdf}
% }
% \subfloat[]{%
%   \includegraphics[width=\columnwidth]{images/ngramDependency_1.pdf}
% }

% \subfloat[Fig6b.pdf]{%
%   \includegraphics[clip,width=0.6\columnwidth]{example-image-b}%
% }

% \begin{figure*}[!htb]
% \minipage{0.32\textwidth}
%   \includegraphics[width=\linewidth]{images/ngramDependency_1.pdf}
%   \caption{A really Awesome Image}\label{fig:awesome_image1}
% \endminipage\hfill
% \minipage{0.32\textwidth}
%   \includegraphics[width=\linewidth]{images/ngramDependency_2.pdf}
%   \caption{A really Awesome Image}\label{fig:awesome_image2}
% \endminipage\hfill
% \minipage{0.32\textwidth}%
%   \includegraphics[width=\linewidth]{images/ngramDependency_3.pdf}
%   \caption{A really Awesome Image}\label{fig:awesome_image3}
% \endminipage
% \end{figure*}

\subsection{Comparison with LanguageTool}
We compared the performance of system with one of the most popular rule-based systems, LanguageTool (LT). Due to some license issues, we could only run LT for 11 languages \textit{viz.,} Danish, Dutch, French, German, Greek, Polish, Portuguese, Romanian, Russian, Spanish and Swedish.
% We created a dataset of 10,000 subtitles each where our detection system flagged exactly one error. 

\begin{figure}[h]
    \includegraphics[width=\columnwidth]{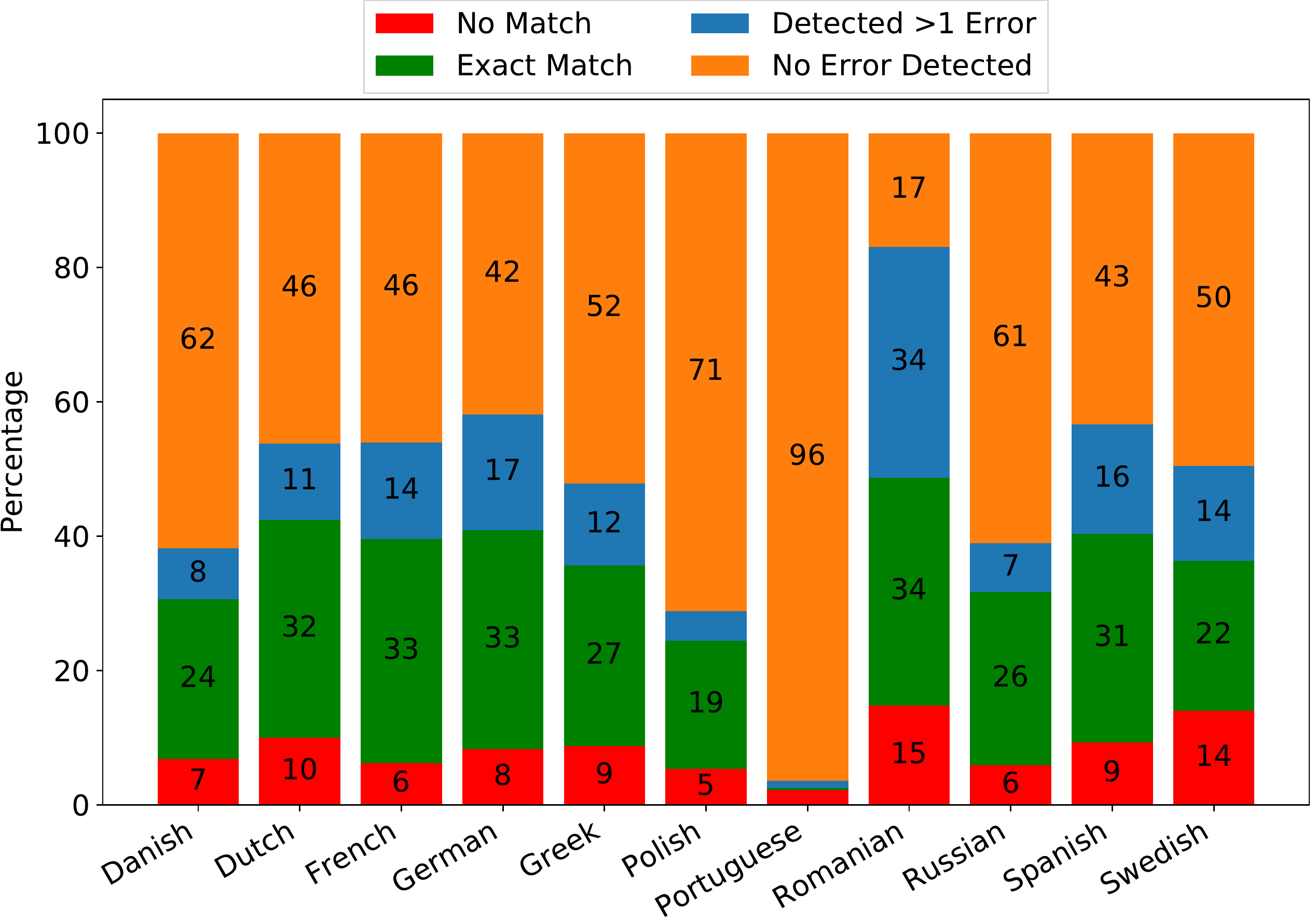}
    \caption{Performance Comparison with LT for 11 languages
    % for Danish, Dutch, French, German, Greek, Polish, Portuguese, Romanian, Russian, Spanish and Swedish
    }
    \label{fig:languageToolComparison}
\end{figure}

As shown in Figure \ref{fig:languageToolComparison}, LT doesn't detect any error in many cases. For example --- for German, it did not detect any error in 42\% sentences and for 25\% (8\% (No Match) + 17\% (Detected more than one error)), it detected more than one error in a sentence out of which in 8\% sentences, the error detected by our system was not detected by LT. Only for 33\% sentences LT detected exactly one error which was same as detected by our system. Results for Portuguese seem very skewed which can be due to the fact Portuguese has two major versions, Brazilian Portuguese (pt-BR) and European Portuguese (pt-PT); LT has different set of rules for both versions whereas dataset used was a mix of both.

\subsection{Public Datasets results}
\label{section:publicdatasetsresults}
\begin{figure*}[t]
\centering
\subfigure[]{
\includegraphics[width=.45\linewidth]{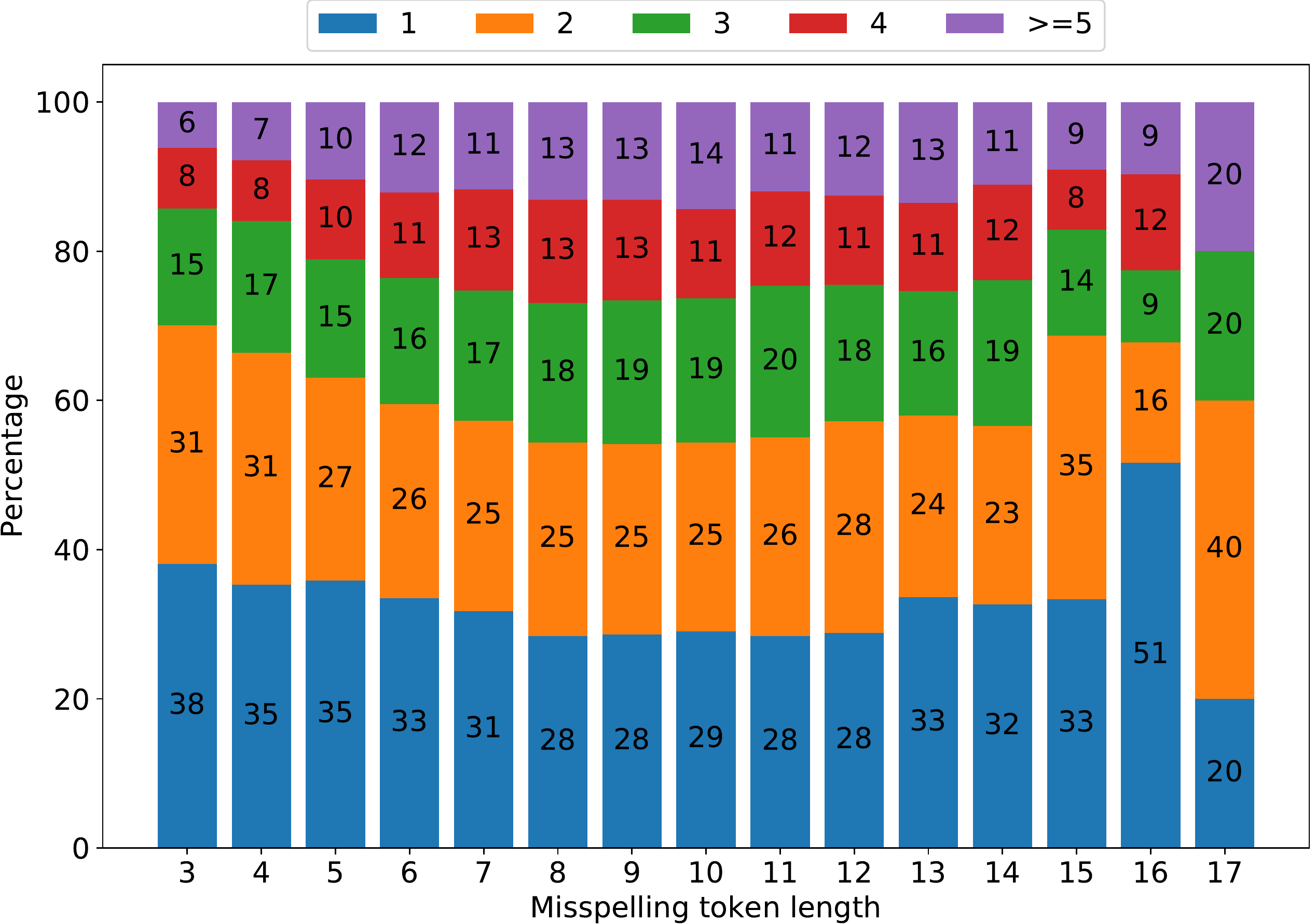}
\label{fig:EditdistancedistributionforPublicEnglishDatasets}
}
\subfigure[without \textbf{birbeck}]{
\includegraphics[width=.45\linewidth]{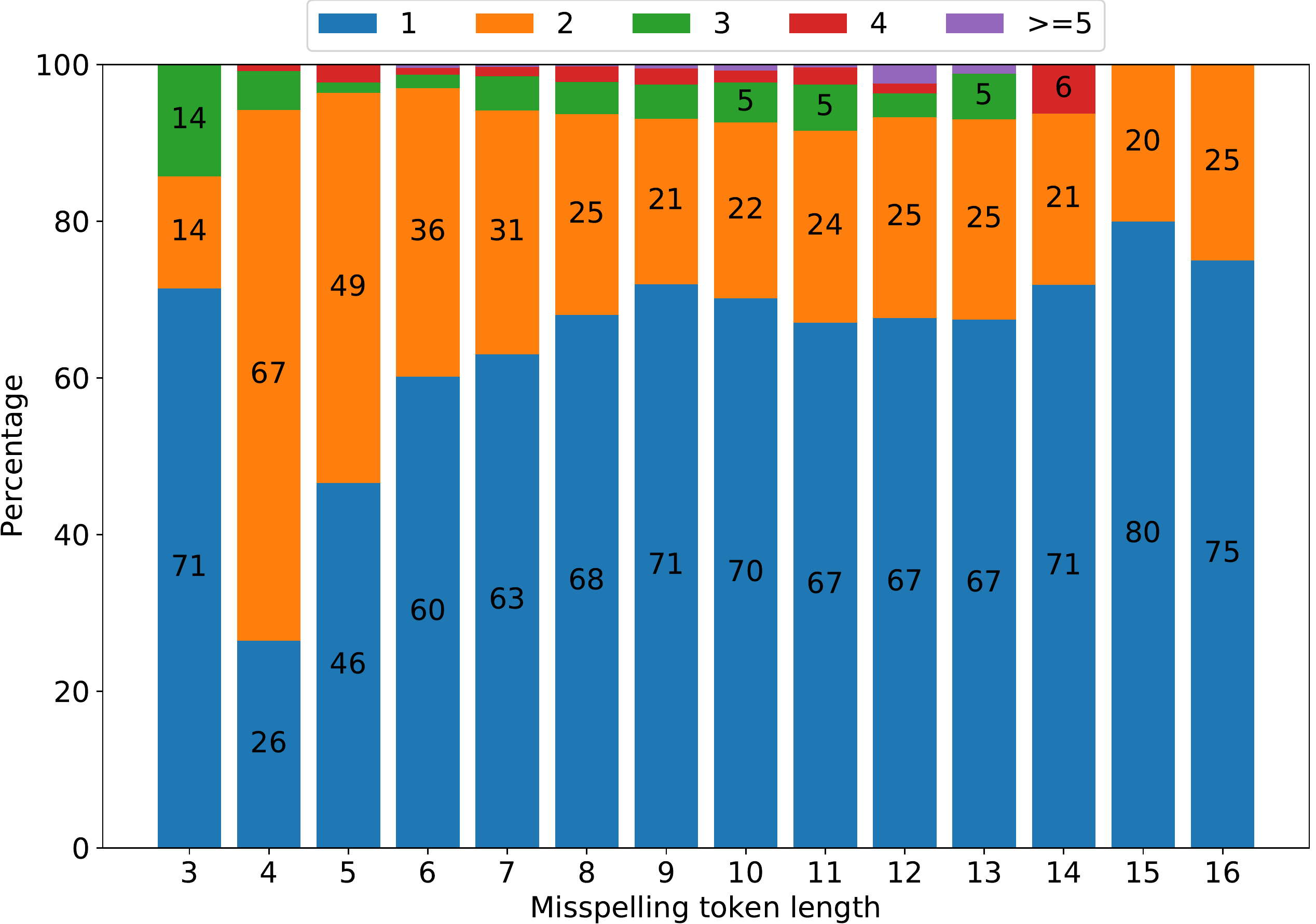}
\label{fig:EditdistancedistributionforPublicEnglishDatasetsWikipedia}
}
\caption{Edit distance distribution for Public English Datasets}
% \label{fig:ngramDependency}
\end{figure*}

We used four publicly available datasets for English --- \textbf{birkbeck}: contains errors from Birkbeck spelling error corpus\footnote{\url{http://ota.ox.ac.uk/}}, \textbf{hollbrook}: contains spelling errors extracted from passages in book, \textit{English for the Rejected}, \textbf{aspell}: errors collected to test GNU Aspell\footnote{\url{http://aspell.net/}} \cite{Deorowicz2005CORRECTINGSE}, \textbf{wikipedia}: most common spelling errors on Wikipedia. Each dataset had a list of misspelling and the corresponding correction. We ignored all the entries which had more than one tokens. We extracted 5,987 unique correct words and 31,589 misspellings. Figure \ref{fig:EditdistancedistributionforPublicEnglishDatasets} shows the distribution of edit distance between misspelling and its correction. Figure \ref{fig:EditdistancedistributionforPublicEnglishDatasetsWikipedia} shows the same distribution excluding \textbf{birkbeck} dataset leaving 2,081 unique words and 2,725 misspellings. \textbf{birkbeck} dataset is the biggest out of four but the quality of this dataset is questionable. As explained by the dataset owners, the dataset is created using poor resources. From Figure \ref{fig:EditdistancedistributionforPublicEnglishDatasetsWikipedia}, our assumption of most of the common misspelling being in maximum edit-distance of 2 is correct. 

% \begin{figure}[h]
%     \includegraphics[width=\linewidth]{images/knownDatasetsDistribution.pdf}
%     \caption{Edit distance distribution for Public English Datasets}
%     \label{fig:EditdistancedistributionforPublicEnglishDatasets}
% \end{figure}

% \begin{figure}[h]
% \centering
%     \includegraphics[width=0.8\linewidth]{images/knownDatasetsDistributionWikipedia.pdf}
%     \caption{Edit distance distribution for Public English Datasets (without \textbf{birkbeck})}
%     \label{fig:EditdistancedistributionforPublicEnglishDatasetsWikipedia}
% \end{figure}

\begin{table}[h]
  \centering
  \caption{Public dataset comparison results}
  \label{table:existingSystems}
%   \resizebox{0.6\columnwidth}{!}{%
    \begin{tabular}{rrrrrr}
    \centering
        & \textbf{P@1} & \textbf{P@3} & \textbf{P@5} & \textbf{P@10} \\ \hline
        Aspell & 60.82 & 80.81 & 87.26 & 91.35 \\
        Hunspell & 61.34 & 77.86 & 83.47 & 87.04 \\
        \textit{Ours} & 68.99 & 83.43 & 87.03 & 90.16 \\
    \end{tabular}
    % }
\end{table}

We use every correct and incorrect token in this dataset to check if they are present and absent in our dictionary respectively in order to prove if our detection system is able to detect correctness/incorrectness of tokens efficiently. The detection system was able to detect $99.13\%$ of correct tokens  and $88.37\%$ of incorrect tokens accurately. The percentage of incorrect token detection is comparatively low is because there are many tokens in dataset which were actually correct but were added in misspelling dataset --- \textit{``flower''}, \textit{``representative''}, \textit{``mysteries''}, etc. Some correct words in dataset which were detected incorrect were also noise due to the fact some words start with a capital letter but in dataset they were in lowercase --- \textit{``beverley''}, \textit{``philippines''}, \textit{``wednesday''} etc. Comparison of most popular spell checkers for English (GNU Aspell and Hunspell\footnote{\url{http://hunspell.github.io/}}) on this data is presented in Table \ref{table:existingSystems}. Since these tools only work on word-error level, we used only unigram probabilities for ranking. Our system outperforms both the systems.

% \textbf{\textit{holbrook} dataset}: contains English sentences with marked misspelling and their corrections. We removed all the errors where the incorrect word or correct word were tokenized into more than one tokens. We also removed all real-word errors. The final dataset was small having 446 sentences with a total of 998 errors. Results are reported in Table \ref{table:holbrookdataresults}. Dataset contained some errors where the edit distance was greater than 2, therefore \textit{P@Inf} is not 100\%.

% \begin{table}[h]
%   \centering
%   \caption{\textit{holbrook} dataset Performance results}
%   \label{table:holbrookdataresults}
% %   \resizebox{0.6\columnwidth}{!}{%
%     \begin{tabular}{rrrrr}
%     \centering
%         \textbf{P@1} & \textbf{P@3} & \textbf{P@5} & \textbf{P@10} & \textbf{P@Inf} \\ \hline
%         44.23 & 62.86 & 71.15 & 80.29 & 87.01 \\
%     \end{tabular}
%     % }
% \end{table}
% \begin{table}[h]
%     \caption{Caption}
%     \label{tab:my_label}
%     \begin{tabular}{l|l|c|c|c}
%         \multicolumn{2}{c}{}&\multicolumn{2}{c}{True diagnosis}&\\
%         \cline{3-4}
%         \multicolumn{2}{c|}{} & Positive & Negative & \multicolumn{1}{c}{Total}\\
%         \cline{2-4}
%         \multirow{2}{*}{Detected} & Positive & $7446$ & $4109$ & $a+b$\\
%         \cline{2-4}
%         & Negative & $65$ & $31094$ & $c+d$\\
%         \cline{2-4}
%         \multicolumn{1}{c}{} & \multicolumn{1}{c}{Total} & \multicolumn{1}{c}{$a+c$} & \multicolumn{    1}{c}{$b+d$} & \multicolumn{1}{c}{$N$}\\
%     \end{tabular}
% \end{table}

\subsection{False Positive evaluation}
For a spell checker system, false positives is when spelling error is detected but there was none. We experimented with a mix of three public datasets --- \textbf{OpenSubtitles dataset}\cite{Lison2016OpenSubtitles2016EL}, \textbf{OPUS Books dataset}\cite{TIEDEMANN12.463} and \textbf{OPUS Tatoeba dataset}\cite{TIEDEMANN12.463} to generate a dataset with minimum 15,000 words for each of 24 languages. Since these datasets are human curated, we can safely assume every token should be detected as a \textit{known} word.

As shown in Table \ref{table:falsepositiveresults}, most of the words for each language were detected as known but still there was a minor percentage of words which were detected as errors. For English, the most frequent errors in complete corpus were either proper nouns or foreign language words --- \textit{``Pencroft''}, \textit{``Oblonsky''}, \textit{``Spilett''}, \textit{``Meaulnes''} and \textit{``taient''}. This proves the effectiveness of system against false positives. 
\begin{table}[h]
  \centering
  \caption{False Positive Experiment Results}
  \label{table:falsepositiveresults}
  \resizebox{\columnwidth}{!}{%
    \begin{tabular}{lcrrrc}
    \centering
    \textbf{Language} & \textbf{\# Sentences} & \textbf{\# Total Words} & \textbf{\# Detected} & \textbf{\%} \\ \hline
        Bengali & 663748 & 457140 & 443650 & 97.05 \\
        Czech & 6128 & 36846 & 36072 & 97.90 \\
        Danish & 16198 & 102883 & 101798 & 98.95 \\
        Dutch & 55125 & 1048256 & 1004274 & 95.80 \\
        English & 239555 & 4981604 & 4907733 & 98.52 \\
        Finnish & 3757 & 43457 & 39989 & 92.02 \\
        French & 164916 & 3244367 & 3187587 & 98.25 \\
        German & 71025 & 1283239 & 1250232 & 97.43 \\
        Greek & 1586 & 43035 & 42086 & 97.79 \\
        Hebrew & 95813 & 505335 & 494481 & 97.85 \\
        Hindi & 5089 & 37617 & 37183 & 98.85 \\
        Indonesian & 100248 & 84347 & 82809 & 98.18 \\
        Italian & 36026 & 718774 & 703514 & 97.88 \\
        Marathi & 17007 & 84286 & 79866 & 94.76 \\
        Polish & 3283 & 34226 & 32780 & 95.78 \\
        Portuguese & 1453 & 25568 & 25455 & 99.56 \\
        Romanian & 4786 & 34862 & 34091 & 97.79 \\
        Russian & 27252 & 384262 & 372979 & 97.06 \\
        Spanish & 108017 & 2057481 & 2028951 & 98.61 \\
        Swedish & 3209 & 66191 & 64649 & 97.67 \\
        Tamil & 40165 & 21044 & 19526 & 92.79 \\
        Telugu & 30466 & 17710 & 17108 & 96.60 \\
        Thai & 16032 & 67507 & 49744 & 73.69 \\
        Turkish & 163910 & 794098 & 775776 & 97.69 \\

    \end{tabular}
    }
\end{table}

\section{Conclusion}
\label{section:conclusion}
We presented a novel context sensitive spell checker system which works in real-time. Most of the available literature majorly discuss spell checkers for English and sometimes for some European (like German, French) and Indian languages (like Hindi, Marathi), but there are no publicly available systems (non-rule based) which can work for all languages.

Our proposed system outperformed industry-wide accepted spell checkers (GNU Aspell and Hunspell) and rule-based spell checkers (LanguageTool). First, we proposed three different approaches to create typographic errors for any language which has not been done earlier in multilingual setting. Second, we divide our proposed system in 5 steps --- Preprocessing; tokenization; error detection; candidate suggestion generation; and suggestion ranking. We used n-gram conditional probability dictionaries to understand context to rank suggestions and present top suggestions.

% We use a custom tokenizer which can be easily extended to new languages with minimal language expertise. For detection, we use a dictionary of words generated using Wikipedia articles data and manually curated subtitle data to compensate for lack of Wikipedia data for many languages. 
We showed the adaptability of our system to 24 languages using precision@k for $k \in {1, 3, 5, 10}$ and mean reciprocal rank (MRR). The system performs at a minimum of 80\% \textit{P@1} and 98\% \textit{P@10} on synthetic dataset. We showed the robustness of our system to false-positives. In future, we can further increase the support to real-word errors and compound word errors.
% Future endeavours:

% \section{Future Work / Possible Improvements}
% \label{section:futurework}
% \begin{enumerate}
    % \item \textbf{Extending to new languages}: In future, we can increase the language coverage of the system to cover more resource-scarce languages like Vietnamese, Punjabi and Malay. It would be interesting to see the performance of system straight-out-of-box for languages where segmentation and tokenization is very different like Arabic \cite{Monroe2014WordSO}, Chinese, Japanese and Korean. 
    % \item \textbf{Reduce dictionary size}: Even though we reduced the size of dictionaries significantly but still the size of dictionaries is huge for any system to load in memory. A distributed database solution (like Berkeley DB \cite{Olson:1999:BD:1268708.1268751}, Memcached \cite{Fitzpatrick2004DistributedCW} or Redis\footnote{\url{https://redis.io}}) can be used to make the system more adoptable in real situations.
    % \item \textbf{Extending to more error categories}: 
    % \item \textbf{Experiment with neural word embeddings for ranking suggestions}: Word embeddings \cite{Mikolov2013DistributedRO, DBLP:journals/corr/BojanowskiGJM16}, theoretically, should be able to overcome the problem of unknown n-grams and provide a better and cheaper contextual information than n-grams. We performed some preliminary experiments with weighted sum of cosine distance of all the tokens in a context window with each suggestion and ranking based on that score. The results were not very promising but with more extensive experimentation, results can be improved. 
    % \item front end tool
% \end{enumerate}

\bibliographystyle{IEEEtran}
\bibliography{main}

\end{document}